\documentclass[runningheads,a4paper]{llncs}
\usepackage{tabularx}
\usepackage{hyperref}
\usepackage{amssymb}
\usepackage{graphicx,epstopdf,epsfig,subcaption,amsmath}
\usepackage{setspace}
\usepackage{hhline}
\usepackage{url}
\usepackage{multirow}
\usepackage{array,booktabs}% http://ctan.org/pkg/{array,booktabs} 

\urldef{\mailsa}\path|{alfred.hofmann, ursula.barth, ingrid.haas, frank.holzwarth,|
\urldef{\mailsb}\path|anna.kramer, leonie.kunz, christine.reiss, nicole.sator,|
\urldef{\mailsc}\path|erika.siebert-cole, peter.strasser, lncs}@springer.com|
\newcommand{\keywords}[1]{\par\addvspace\baselineskip
\noindent\keywordname\enspace\ignorespaces#1}
\usepackage{floatrow}
\newfloatcommand{capbtabbox}{table}[][\FBwidth]

\newsavebox\CBox
\def\textBF#1{\sbox\CBox{#1}\resizebox{\wd\CBox}{\ht\CBox}{\textbf{#1}}}
\begin{document}
	
\mainmatter  % start of an individual contribution

\newcommand{\etal}{\textit{et al}.}
\newcommand{\ie}{\textit{i}.\textit{e}.}
\newcommand{\eg}{\textit{e}.\textit{g}.}

\title{End-to-End Adversarial Shape Learning for Abdomen Organ Deep Segmentation}

\author{
	Jinzheng Cai\inst{1}, Yingda Xia\inst{2}, Dong Yang\inst{2}, 
	\\
	Daguang Xu\inst{2}, Lin Yang\inst{1}, Holger Roth\inst{2}
	\\
	\institute{
	University of Florida, Gainesville, FL, 32611, USA
 	\and NVIDIA, Bethesda, MD, 20814, USA
	\\
	\email{caijinzhengcn@gmail.com, hroth@nvidia.com}}
}
% \author{Submission id: 38}

\titlerunning{Adversarial Shape Learning for Abdomen Organ Segmentation}
\authorrunning{J. Cai~\etal{}}
% \authorrunning{***}

% \author{Anonymous Submission to MICCAI}
%
% \authorrunning{Lecture Notes in Computer Science: Authors' Instructions}

% \institute{ID: }


\toctitle{Lecture Notes in Computer Science}
\tocauthor{Authors' Instructions}
\maketitle

\begin{abstract}

Automatic segmentation of abdomen organs using medical imaging has many potential applications in clinical workflows. Recently, the state-of-the-art performance for organ segmentation has been achieved by deep learning models, \ie{}, convolutional neural network (CNN). However, it is challenging to train the conventional CNN-based segmentation models that aware of the shape and topology of organs. In this work, we tackle this problem by introducing a novel end-to-end shape learning architecture -- organ point-network. It takes deep learning features as inputs and generates organ shape representations as points that located on organ surface. We later present a novel adversarial shape learning objective function to optimize the point-network to capture shape information better. We train the point-network together with a CNN-based segmentation model in a multi-task fashion so that the shared network parameters can benefit from both shape learning and segmentation tasks. We demonstrate our method with three challenging abdomen organs including liver, spleen, and pancreas. The point-network generates surface points with fine-grained details and it is found critical for improving organ segmentation. Consequently, the deep segmentation model is improved by the introduced shape learning as significantly better Dice scores are observed for spleen and pancreas segmentation.


\end{abstract}

\section{Introduction}
\label{secion:introduction}

Automatic organ segmentation is becoming an increasingly important technique providing supports for routine clinical diagnoses and treatment plannings. Organ segmentation in medical images, \eg{}, computed tomography (CT) and magnetic resonance imaging (MRI), is usually formulated as a voxel-wise classification problem. Recently, deep learning methods (\eg{}, fully convolutional network (FCN) based segmentation~\cite{DBLP:conf/miccai/RonnebergerFB15,DBLP:conf/miccai/CicekALBR16,DBLP:journals/mia/RothLLHFSS18}) have been reported as powerful baselines to various segmentation tasks, where the deep learning methods can perform reliably on the majority cases. However, segmentation error often occur near the organ surface largely due to low image quality, vague organ boundaries, and large organ shape variation. Although several attempts \cite{j_cai_miccai17_pancreas,DBLP:conf/miccai/YangXZGCGMC17} have been reported in the literature, it is still challenging for deep learning models to produce segmented results with smooth and realistic shapes as it would require strong global reasoning ability to model relations between all image voxels. 

\par \qquad 
We propose a shape learning network -- organ point-network to improve the segmentation performance of FCN-based methods. The organ point-network takes deep learning features as its inputs and the deep learning features are extracted from raw 3D medical images by 3D convolutional neural network layers. The point-network outputs shape representations of target organs, where the shape representations are defined as sets of points locating on organ surface. The idea of point-network is inspired by a point set generator that proposed by Fan \etal{} \cite{DBLP:conf/cvpr/FanSG17}. In \cite{DBLP:conf/cvpr/FanSG17}, the point generator is built with 2D convolutional neural network layers. It takes 2D images and segmentation masks as its inputs and outputs sets of points as 3D reconstructions of target objects. To obtain 3D reconstruction from 2D images, it requires to introduce moderate level of uncertainty. Thus, the point generator in \cite{DBLP:conf/cvpr/FanSG17} uses only coarse-scale features from the top network layer. Therefore, it generates points that lacks of fine-grained shape details. Differently, the 3D organ segmentation setup requires our organ point-network to process 3D information and to use only the deep features due to the lack of available segmentation masks. Besides, the organ point-network needs to reconstruct organ accurately with recovering as may shape details as possible, otherwise, the segmentation model could be significantly distracted by the inaccurate shape information under the multi-task learning context.

\par \qquad 
In this work, we introduce shape learning to improve 3D FCN-based organ segmentation. We summarize our contributions as, 1) we evaluate the multi-task learning mechanism, which jointly optimizes the segmentation task with the shape learning task and improves the segmentation results; 2) we propose a novel organ point-network and a effective adversarial learning strategy for accurate organ reconstruction; 3) we demonstrate our method with extensive experiments and report significant segmentation improvements.

\section{Method} 
\label{section:method}

\subsection{Multi-task Learning}

\begin{figure}[t!]
	\includegraphics[width=0.98\linewidth]{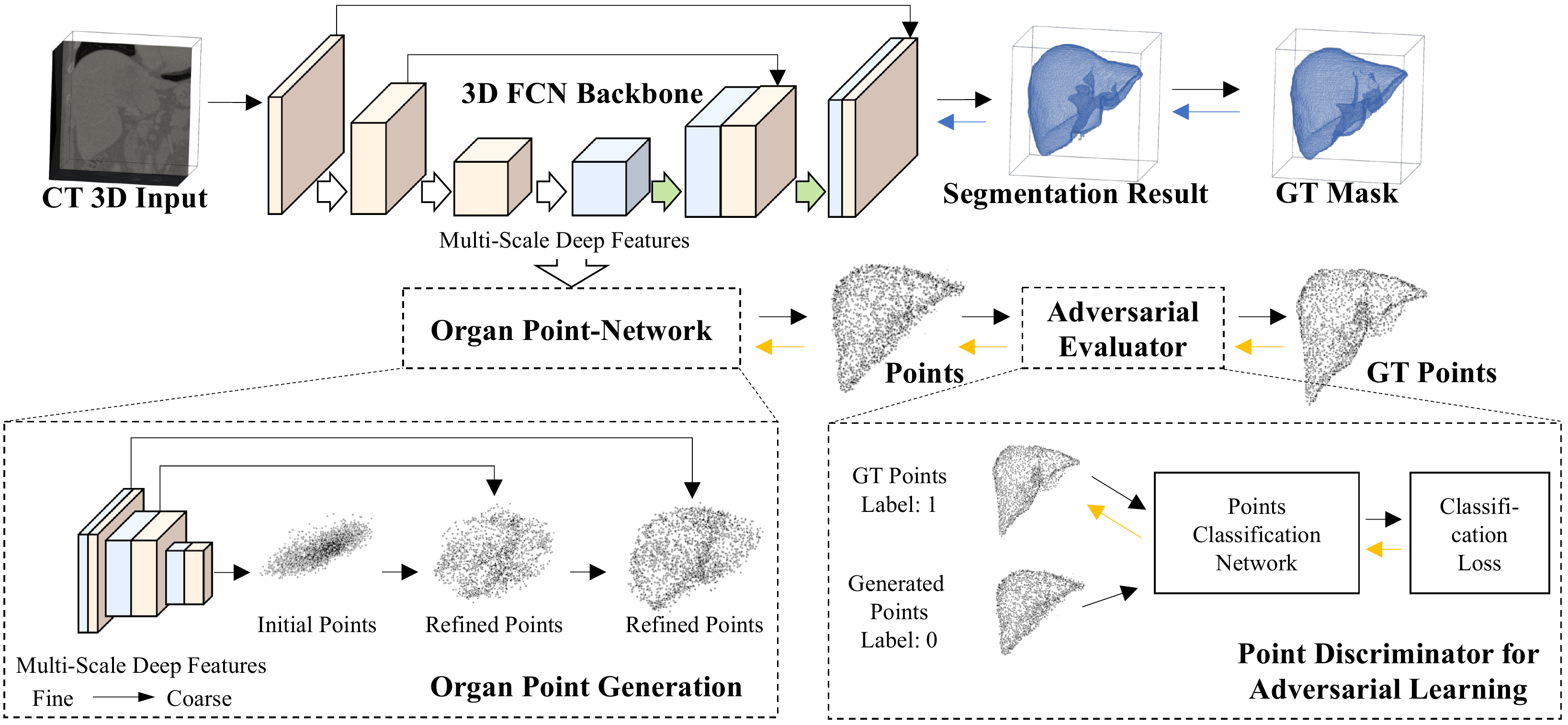}
	\caption{Pipeline of the proposed multi-task learning.}
	\label{fig:pipe-line} 
\end{figure}

In our multi-task learning setup, the proposed organ point-network is jointly optimized with FCN-base segmentation model. Fig. \ref{fig:pipe-line} shows the pipeline of the proposed multi-task learning. The pipeline consists of a 3D FCN backbone, a segmentation loss branch, and a shape learning branch. Specifically, the 3D FCN backbone is built with 3D convolutional layers and it extracts deep features from 3D CT inputs. Resolution of the deep features ranges from coarse to fine, where the coarse-scale feature contains more high-level representations and less image details, while the fine-scale feature has more image details but less high-level representations. In the pipeline, the segmentation loss branch takes the fine-scale feature into a Sigmoid activation layer and outputs the segmentation results. The shape learning branch contains the proposed organ point-network and adversarial evaluator. It takes the multi-scale deep features and outputs organ surface points. 

\subsection{Organ Point-network}
\label{subsec:c2f-s-p-g}

\begin{figure}[t!]
	\includegraphics[width=.98\linewidth]{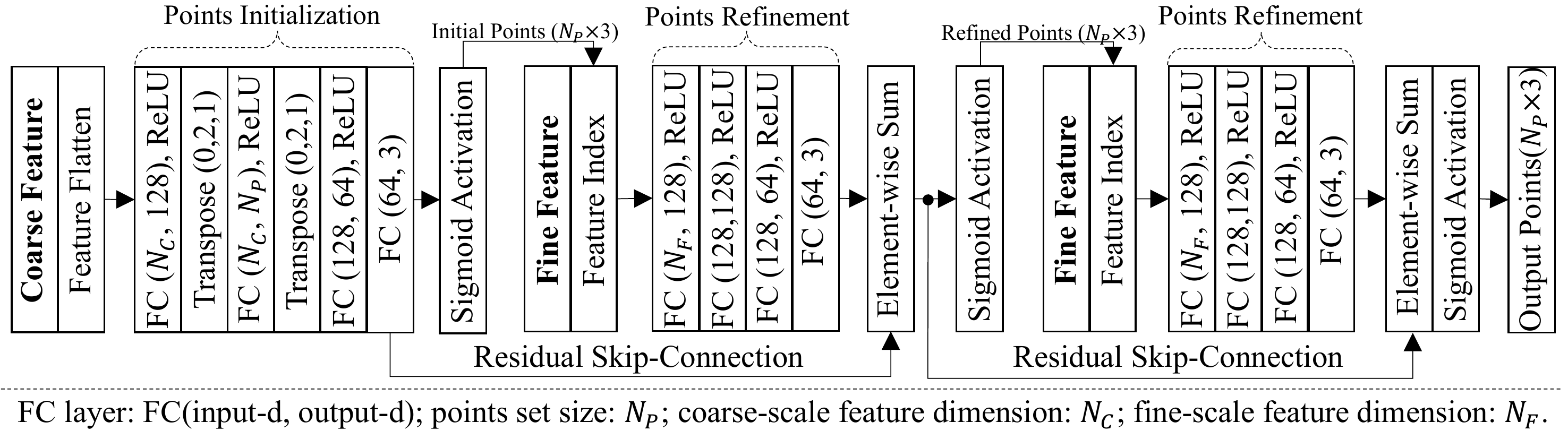}
	\caption{Architecture configuration of the proposed organ point-network.}
	\label{fig:point-network} 
\end{figure}

We design the organ point-network to take multi-scale deep features as its inputs. It first initializes points using the coarsest-scale feature and then fine-tunes the location of each point using the fine-scale features to factor in local shape information. Fig. \ref{fig:point-network} shows the network configuration of the proposed organ point-network. To name the network components, we denote the coarse-scale feature input as $X_C$, and the fine-scale feature input as $X_F$. We formulate the points initialization procedure as a mapping function $\mathcal{F}_{i}(\cdot; \theta_{i})$, where $\theta_{init}$ is the set of network parameters. Similarly, we define the first and second points refinement procedures as $\mathcal{F}_{r1}(\cdot; \theta_{r1})$, and $\mathcal{F}_{r2}(\cdot; \theta_{r2})$, respectively. We propose a ``feature index'' network layer and refer to it as $\mathcal{M}_{FI}(\cdot)$.  For presentation clarity, we omit the standard network components,  such as ReLU, Sigmoid activation, Batch-Normalization, and element-wise summation layers. The point set is presented as a $N_P$ by 3 matrix, where $N_P$ presents the number of points and 3 is the dimension. The point size $N_P$ is fixed along the initial and refined point sets.

\par \qquad 
The points $P \in R^{N_P \times 3}$ is initialized as
\begin{equation}
	P = \mathcal{F}_{i}(X_C; \theta_i), 
\end{equation}
where the coarse-scale feature $X_C$ is from the top layer of the 3D FCN backbone. As passed through multiple 3D convolution and spatial pooling layers, each voxel in $X_C$ should represent the aggregated information of a sub-region from the input image. In other words, the coarse-scale feature $X_C$ contains high-level image representations of the image but lacks of local details. Thus, points in $P$ present only a rough shape of the target organ lacking of details. Fig. \ref{fig:pipe-line} shows an example of the initial points. 

\par \qquad 
Later, the point-network refines each point in $P$ using the fine-scaled feature and this procedure is referred as ``points refinement'' in Fig. \ref{fig:point-network}. A ``feature index'' layer $M_{FI}(\cdot)$ is defined to extract point-wise local feature, 
\begin{equation} \label{eq:fi}
	f_i = M_{FI}(X_F, p_i), 
\end{equation}
where $p_i \in R^{1 \times 3}$ is the i-th point in $P$, and $f_i$ is the indexed local feature. The three coordinates in $p_i$ are normalized values in the range $[0,1]$. Thus, $M_{FI}$ first scale the three coordinates in $p_i$ with the width, height, and depth of $X_F$, respectively. Then, $M_{FI}$ extracts the 3$\times$3$\times$3 sub-region in $X_F$ that centered at the scaled $p_i$ location.  The extracted 3$\times$3$\times$3 sub-region contains local image information at position $p_i$ and it is flattened as $f_i$. Like standard pooling layers, the ``feature index'' layer processes only indexing operations and no network parameter need to be learned, thus the point-network with ``feature index'' layers can be trained end-to-end.

\par \qquad 
Given the fact that the indexed feature $f_i$ contains only local image information, learning of local point movement would be inherently easier than predicting the global coordinate. Thus to refine point $p_i$, we formulate the refinement as a residual learning
\begin{equation} \label{eq:res}
	p_i = p_i + \mathcal{F}_{r1}(f_i; \theta_{r1}). 
\end{equation}
It is implemented via a ``Skip-Connection'' between the points initialization module and the points refinement module (see Fig. \ref{fig:point-network}). Thus, the ``points refinement'' is a combination of Eq. \ref{eq:fi} and Eq. \ref{eq:res} and its matrix form is
\begin{equation} 
	P = P + \mathcal{F}_{r1}(M_{FI}(X_F, P); \theta_{r1}).
\end{equation}

\par \qquad 
Practically, some of the initial points would locate close to the actual organ surface while the others could be far away. Therefore, a second ``points refinement'' is required in the point-network to secure the convergence. In the same spirit of the first refinement, we formulate the second refinement as
\begin{equation}
	P = P + \mathcal{F}_{r2}(M_{FI}(X_F, P); \theta_{r2}).
\end{equation}
There is a ``Skip-Connection'' between the first and second refinements. 

\subsection{Adversarial Evaluator}
\label{subsec:p-c-s} 

In this section, we discuss point-based training objectives to train the proposed point-network. In \cite{DBLP:conf/cvpr/FanSG17}, Fan \etal{} uses Chamfer distance (CD), Earth Mover's distance (EMD) for shape learning. These objectives first find one-versus-one point correspondence between the predicted points $P$ and the ground-truth points $P_{gt}$ (where $P_{gt} \in R^{N_p \times 3}$). Then, difference between the paired predicted point and group-truth point is calculated and the mean difference of the $N_P$ point pairs is used as prediction error that back-propagated to update the network parameters. Due to the calculation of mean differences, these objectives are not sensitive to outliers, which lie far away from the organ surface. Thus, we further propose a novel adversarial learning (AL) loss that works complementary to the CD and EMD metrics to remove the outlier points. During model training, the proposed adversarial loss can be jointly optimized with CD, and EMD losses.

\par
\begin{figure}[t!]
	\includegraphics[width=.98\linewidth]{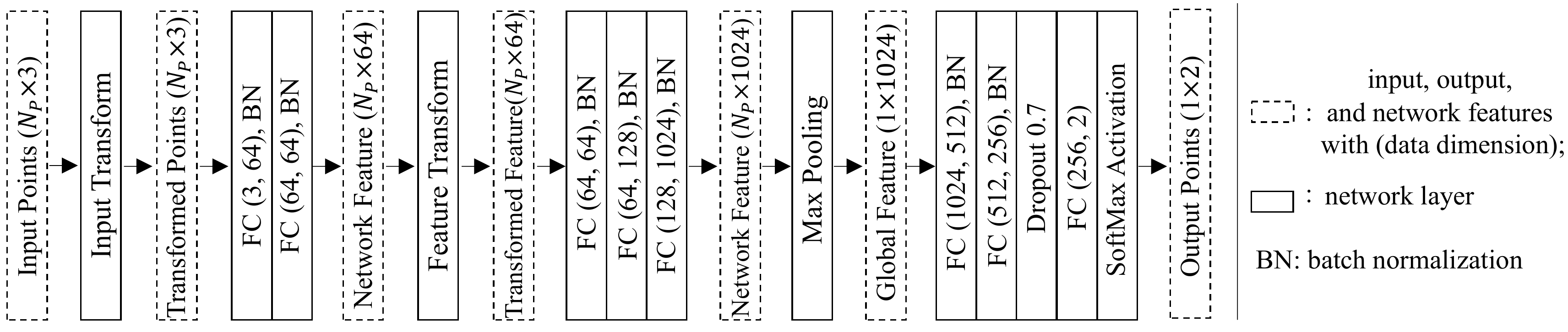}
	\caption{Architecture configuration of the points classification network.}
	\label{fig:pointnet} 
\end{figure}

 \qquad
In the adversarial evaluator, a point set classifier is trained to differentiate the generated points $P$ and ground truth points $P_{gt}$. The classifier would project the points onto the manifold of the target anatomy. Fig. \ref{fig:pointnet} shows the architecture configuration of the points classification network. The classifier takes $N_P$ by 3 points matrix as input and outputs its tag of the input point set. In the classifier, we use ``transformation'' layers proposed in \cite{DBLP:conf/cvpr/QiSMG17}, so that the learned representation by the point set in invariant to geometric transformations. We define the classifier as $\mathcal{D}(\cdot; \theta_D)$, and labels for the generated points $P$ and ground truth points $P_{gt}$ are 0 and 1, respectively. Then, the loss function for adversarial learning is,  
\begin{equation}
\mathcal{L}_{AL} = H(\mathcal{D}(P; \theta_D), 0) + H(\mathcal{D}(\hat{P}_{gt}; \theta_D), 1), 
\end{equation} 
where $H(\cdot,\cdot)$ is the cross entropy loss function, and $\hat{P}_{gt}$ are the ground truth points $P_{gt}$ with randomly added noise that generated from a specified range (\ie{} $[-0.005,0.005]$ in our experiments). The noise is applied to balance the classifier's convergence speed over $P$ and $P_{gt}$. 

\if 0
\par \qquad
\textbf{Jointly optimize multiple supervisions:} The overall objective for training of image-to-point translation is,
\begin{equation} \label{eq1}
\mathcal{L} = \sum_{\mathcal{L}_i \in \mathcal{S}} \lambda_i \mathcal{L}_i + \mathcal{L}_{seg},
\end{equation}
where $S{}={}\{\mathcal{L}_{cd}, \mathcal{L}_{emd}, \mathcal{L}_{rep}, \mathcal{L}_{adv}\}$ is the set of evaluation metrics, and $\lambda_i$ are scale factors to balance the losses. We also use a cross-entropy segmentation loss $\mathcal{L}_{seg}$ to supervise the CNN base $\mathcal{F}_{cnn}$ that generates image feature $X$. In model training, $\mathcal{L}_{seg}$,  $\mathcal{L}_{cd}$, $\mathcal{L}_{emd}$, $\mathcal{L}_{rep}$, and $\mathcal{L}_{adv}$ values lie in the range from 1e-3 to 1e-1. Thus, we set $\lambda_{cd}$, $\lambda_{emd}$, $\lambda_{rep}$, and $\lambda_{adv}$ to 100, 100, 10, and 1 in our implementation to balance the contribution of each part in the joint loss $\mathcal{L}$. 
\fi 

\section{Experiments and Analyses} 
\label{section:experiments}

% \par \qquad
{\bf Data preparation:} 
To demonstrate the proposed shape learning method, we test three abdomen organs, including liver, pancreas, and spleen. These datasets are publicly available from a organ segmentation competition\footnote{\url{http://medicaldecathlon.com}} and contain 131, 281, and 41 voxel-wise annotated CT volumes for liver, pancreas, and spleen, respectively. For each organ, we randomly separate the images into 50\%, 25\%, and 25\% for training, validating, and testing. To evaluate the effectiveness of the proposed organ point-network, the target organ in each CT volume is center cropped to preserve the whole organ shape information. To include sufficient image background, 20-voxel padding is added to each direction of the cropped volume. Based on the sizes of organs, liver and pancreas volumes are down-sampled to 128$\times$128$\times$128 and spleen volumes are down-sampled 32$\times$128$\times$128. Using the original image resolution for training and testing may obtain better performance, however, we fit the segmentation model, point-network, and points classifier into a single GPU for computation stability, and we mainly focus on the segmentation improvements from the proposed 3D shape learning. 

\par \qquad To generate ground truth points $P_{gt}$, we fuse the organ and lesion annotations to generate the outer organ surface, on which ground truth surface points are generated using the marching cubes algorithm and farthest point sampling~\cite{DBLP:conf/cvpr/FanSG17}. The size of the point set $N_P$ is empirically set to $2048$ in all experiments. The using of $P$ to reconstruct segmentation mask has its own challenges and is out the scope of this paper. Here, we focus on comparing results from segmentation networks that are trained with and without point set generation.

\par \qquad 
{\bf Implementation details:}
We use 3D U-Net \cite{DBLP:conf/miccai/CicekALBR16} as the universal backbone for all segmentation models. It consists of totally 3 times symmetric down-sampling and up-sampling, which are performed by stride 2 max-poling and nearest up-sampling, respectively. Based on the multi-scale features extracted by the 3D U-Net, we construct the proposed organ point-network and point classifier with the architectures as shown in Fig. \ref{fig:point-network} and Fig. \ref{fig:pointnet}, respectively. Network parameters are initialized with Xavier and trained with Adam optimizer. We set the learning rate as 5e-4. The batch-size is fixed to 1 due to the limitation of GPU memory. Since models may converge differently, we train each model with sufficient number of epochs until it gains no performance improvement on the validation set. We then select the model checkpoint with the best validation performance to report the testing result. 
%The number of points $N_P$ is 2048 for all organs. 
All the experiments have been conducted on a NVIDIA TITAN X GPU.

\par \qquad 
We consider the 3D U-Net as a strong 3D FCN backbone. Using 100\% images, the best solution \cite{DBLP:journals/corr/abs-1809-10486} reported in the segmentation competition has reported 95.2\% Dice scores for liver segmentation. In our experiments, only 50\% of the images is applied for model training and the rest is used for validating and testing. However, the 3D U-Net still achieves 94.1\% Dice score.

\begin{table}[t!] 
	\centering
	\small{
	\caption{Evaluation of generated points with EMD, which is shown as mean$\pm$sd.}
	\label{tab:1}
	\begin{tabular}{>{\centering}m{0.85in}>{\centering}m{1.5in}|>{\centering}m{0.75in}>{\centering}m{0.75in}>{\centering\arraybackslash}m{0.75in}}
		\hline
		\multicolumn{2}{c|}{Method} & Liver & Spleen &  Pancreas \\
		\hline
		\multicolumn{2}{c|}{Two-Branch} & 1.66$\pm$1.19 & 4.3$\pm$1.4 & 9.7$\pm$6.8 \\
		\multirow{2}{*}{Point-Network} 
		& w/o Adversarial-Loss & 0.72$\pm$0.44 & 5.2$\pm$2.4 & 9.1$\pm$7.7 \\
		& with Adversarial-Loss & \textBF{0.69$\pm$0.39} & \textBF{3.8$\pm$1.9} & \textBF{8.4$\pm$7.4} \\
		\hline
	\end{tabular}}
\end{table}

\par
\begin{figure}[t!]
	\begin{subfigure}{0.4\textwidth}
		\includegraphics[width=\textwidth]{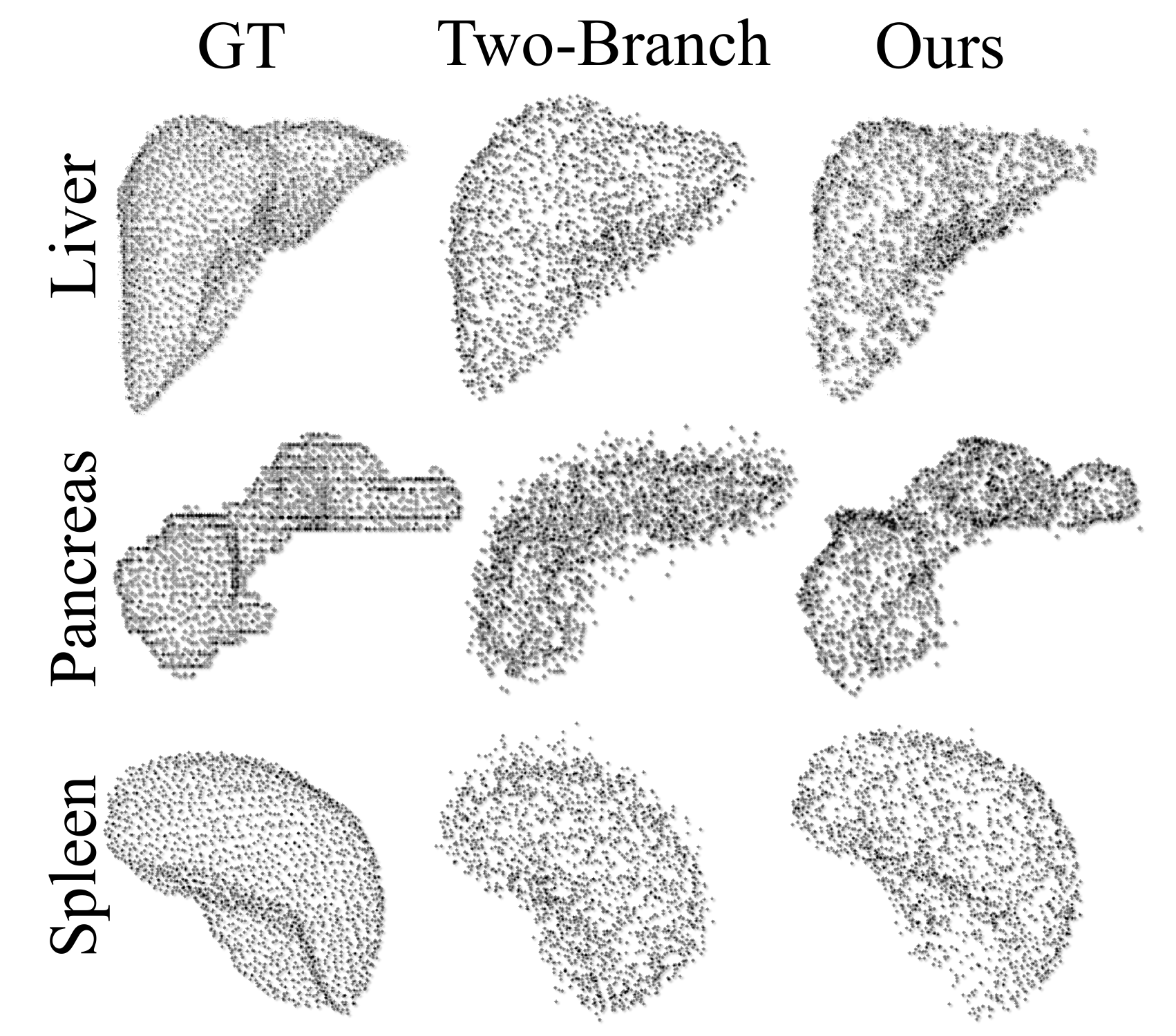}
		\caption{Point-Network vs. Two-Branch}
		\label{fig:3-a}
	\end{subfigure}
	\begin{subfigure}{0.4\textwidth}
		\includegraphics[width=\textwidth]{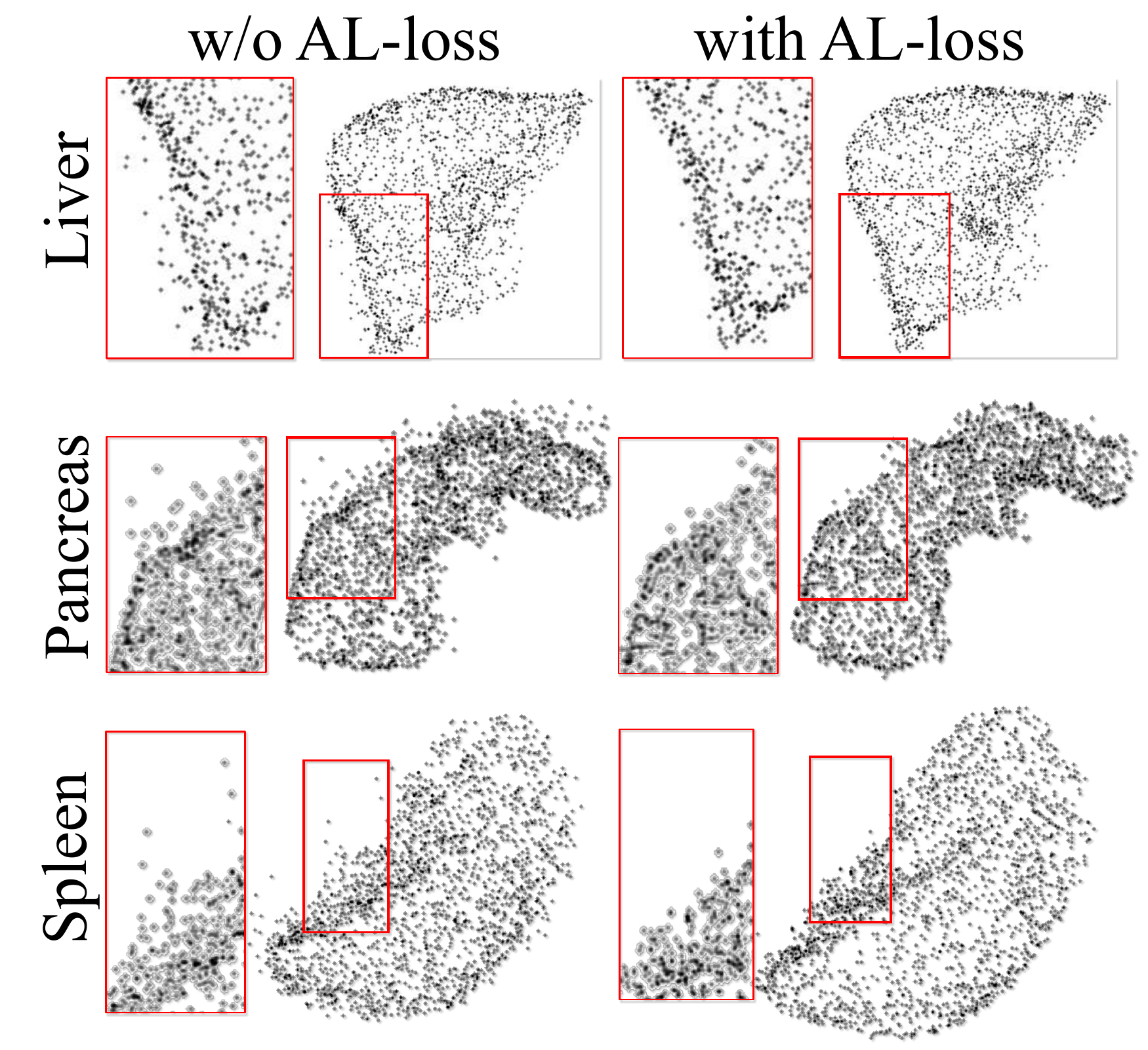}
		\caption{W/o AL-loss vs. AL-loss}
		\label{fig:3-b}
	\end{subfigure}
	\caption{Visualization of generated points.}
	\label{fig:3}
\end{figure}

 \qquad
Firstly, we evaluate the proposed point-network and adversarial learning (AL) loss with quantitative and qualitative analyses. A state-of-the-art point generator -- Two-Branch \cite{DBLP:conf/cvpr/FanSG17} is choose to be the baseline. The Two-Branch model is designed to map 2D images into points using only the global image information. We modify its 2D convolutional network layers with 3D convolutions so that it can process 3D CT volumes. We follow that same setting in \cite{DBLP:conf/cvpr/FanSG17} and normalize coordinates of each point cloud into the range of (0,1) and measure the difference between generated and ground truth points with Earth Mover's distance (EMD). In Table \ref{tab:1}, we present quantitative assessment of different point generation methods. 
We also present visual examples in Fig. \ref{fig:3}. The Two-Branch model performs as a strong baseline outperforming the proposed Point-Network (without AL-loss) for spleen shape learning. However, for organs with more complex surfaces, it is observed that the ``point refinement'' in point-network has improved the generated points with lots of details. The point-network reduces the mean EMD score for liver from 1.66 to 0.72, and for pancreas from 9.7 to 9.1. Fig. \ref{fig:3-a} shows the ground-truth, Two-Branch generated points, and point-network generated points. Comparing point-networks trained with and without AL-loss, we observe systematical improvements. The AL-loss have reduced mean EMD scores for liver, pancreas, and spleen by 4\%, 7\% and 27\%. As shown in Fig. \ref{fig:3-b}, AL-loss has significantly reduced the number of outlier points. 

\par \qquad
In the second experiment, we evaluate segmentation results with Dice using the original CT spacing and report Hausdorff distance (HD), and Average distance (AVGD) \cite{taha2015metrics} in voxel. We compare 3D U-Nets trained with and without point generators, which should introduce shape learning to affect the segmentation performance. Quantitative result shown in Table \ref{tab:2} demonstrate that 3D U-Net can be improved by its attached point generators. The proposed Point-network has significantly\footnote{P-value $<$ 0.05 in Wilcoxon Signed rank test.} improved the Dice scores of spleen and pancreas segmentation. However, we also observe performance degradation when the Two-Branch is applied for liver and spleen segmentation. Based on this observation, we argue that accurate shape learning is critical for improving the segmentation performance, otherwise, it will bring unwanted performance loss. Compared with Two-Branch model, the proposed Point-network generator performs much more stable and robustly improves the 3D U-Net on the three organs in terms of all evaluation metrics.

\begin{table*}[t!]
	 \centering
	 \small{
	 \caption{
         Evaluation of organ segmentation with Dice similarity (DICE), Hausdorff distance (HD), and average distance (AVGD) which are shown in the form of mean$\pm$sd. We use {\bf bold} to indicate experiment sets where 3D U-Net is outperformed and \textbf{\textit{italic bold}} to indicate improvements with statistical significance.
         }
	 \label{tab:2}
	 \begin{tabular}{>{\centering}m{0.8in}>{\centering}m{0.8in}|>{\centering}m{1.0in}>{\centering}m{1.0in}>{\centering\arraybackslash}m{1.0in}}
 		\hline
 		Metric & Organ & 3D U-Net & Two-Branch & Ours \\
 		\hline
 		\multirow{3}{*}{Dice}
 		& Liver & 0.941$\pm$0.034 & 0.925$\pm$0.050 & \textbf{0.944$\pm$0.034} \\
 		& Spleen & 0.955$\pm$0.014 & 0.934$\pm$0.043 & \textbf{\textit{0.960$\pm$0.007}} \\
 		& Pancreas & 0.732$\pm$0.120 & \textbf{\textit{0.734$\pm$0.139}} & \textbf{\textit{0.743$\pm$0.122}} \\
 		\hline
 		\multirow{3}{*}{HD [voxels]}
 		& Liver & 35.862$\pm$17.960 & 39.041$\pm$20.061 & \textbf{31.365$\pm$15.576} \\
 		& Spleen & 6.889$\pm$1.998 & 7.482$\pm$3.454 & \textbf{6.632$\pm$2.543} \\
		& Pancreas & 25.406$\pm$17.953 & \textbf{22.653$\pm$13.109} & \textbf{22.723$\pm$12.536} \\
		\hline
		\multirow{3}{*}{AVGD [voxels]}
		& Liver & 0.325$\pm$0.436 & 0.453$\pm$0.506 & \textbf{0.273$\pm$0.322} \\
		& Spleen & 0.061$\pm$0.033 & 0.101$\pm$0.078 & \textbf{0.051$\pm$0.016} \\
		& Pancreas & 1.080$\pm$1.050 & \textbf{1.074$\pm$1.104} & \textbf{\textit{0.997$\pm$0.991}} \\ 
 		\hline
 	\end{tabular}
	 }
\end{table*}

\section{Conclusion}
\label{section:conclusion}

In this paper, we presented a multi-task deep learning model for shape learning and abdomen organ segmentation. Under the multi-task context, the proposed shape learning model -- point-network uses multi-scale deep learning features from the segmentation model to generate organ surface points with fine-grained details. A novel adversarial learning strategy is introduces to improve the generated points with less outliers. The shape learning then improves the intermediate network layers of the segmentation model and improves organ segmentation. The effectiveness of the proposed method has been demonstrated by experiments of three challenging abdominal organs including liver, spleen, and pancreas.

\bibliographystyle{splncs}
%\bibliography{egbib}

\end{document}